\documentclass{article}

\usepackage{microtype}
\usepackage{graphicx}
\usepackage{subcaption}
\usepackage{booktabs} 
\usepackage{csquotes}

\usepackage{hyperref}

\newcommand{\GC}[1]{{\sf \small \textcolor{magenta}{{\bf GC:} #1}}}

\newcommand{\FP}[1]{{\sf \small \textcolor{orange}{{\bf FP:} #1}}}
\newcommand{\MR}[1]{{\sf \small \textcolor{green}{{\bf MR:} #1}}}
\newcommand{\balpha}{\boldsymbol{\alpha}}
\newcommand{\bw}{\mathbf{w}}

\newcommand{\Ltrain}{\mathcal{L}_{\text{train}}}
\newcommand{\Lval}{\mathcal{L}_{\text{val}}}

\usepackage[preprint]{icml2026}

\usepackage{color}

\usepackage{amsmath}
\usepackage{amssymb}
\usepackage{mathtools}
\usepackage{amsthm}
\usepackage{comment}
\usepackage{multirow}

\usepackage[capitalize,noabbrev]{cleveref}

\theoremstyle{plain}

\theoremstyle{definition}

\theoremstyle{remark}

\usepackage[textsize=tiny]{todonotes}

\icmltitlerunning{Submission and Formatting Instructions for ICML 2026}

\begin{document}

\twocolumn[
  \icmltitle{SQUAD: Scalable Quorum Adaptive Decisions via ensemble of early exit neural networks}

\begin{icmlauthorlist}
  \icmlauthor{Matteo Gambella}{yyy}
  \icmlauthor{Fabrizio Pittorino}{yyy}
  \icmlauthor{Giuliano Casale}{zzz} 
  \icmlauthor{Manuel Roveri}{yyy}
\end{icmlauthorlist}

\icmlaffiliation{yyy}{Dipartimento di Elettronica, Informazione e Bioingegneria (DEIB), Politecnico di Milano, Milan, Italy}
\icmlaffiliation{zzz}{Department of Computing, Imperial College London, London, United Kingdom}

\icmlcorrespondingauthor{Matteo Gambella}{matteo.gambella@polimi.it}

  \icmlkeywords{Machine Learning, ICML}

  \vskip 0.3in
]



\printAffiliationsAndNotice{}  

\begin{abstract} 
Early-exit neural networks have become popular for reducing inference latency by allowing intermediate predictions when sufficient confidence is achieved. However, standard approaches typically rely on single-model confidence thresholds, which are frequently unreliable due to inherent calibration issues. To address this, we introduce SQUAD (Scalable Quorum Adaptive Decisions), the first inference scheme that integrates early-exit mechanisms with distributed ensemble learning, improving uncertainty estimation while reducing the inference time. Unlike traditional methods that depend on individual confidence scores, SQUAD employs a quorum-based stopping criterion on early-exit learners, by collecting intermediate predictions incrementally in order of computational complexity until a consensus is reached and halting the computation at that exit if the consensus is statistically significant. To maximize the efficacy of this voting mechanism, we also introduce QUEST (Quorum Search Technique), a Neural Architecture Search method to select early-exit learners with optimized hierarchical diversity, ensuring learners are complementary at every intermediate layer. This consensus-driven approach yields statistically robust early exits, improving the test accuracy up to 5.95\% compared to state-of-the-art dynamic solutions with a comparable computational cost and reducing the inference latency up to 70.60\% compared to static ensembles while maintaining a good accuracy. 
\end{abstract}

\section{Introduction}

Deep Neural Networks (DNNs) have established themselves as the de facto standard for many applications such as computer vision tasks, natural language processing, audio processing, and time series forecasting \cite{TCN}. However, their remarkable performance typically comes at the cost of massive parameter counts and computational complexity \cite{scalinglaw}. This trade-off presents a critical bottleneck for deployment in resource-constrained environments, such as IoT sensors and Edge computing platforms, where energy budgets are strictly limited and low latency is paramount \cite{alippi2017not}. To mitigate these costs, the research community has increasingly adopted Dynamic Neural Networks for their ability to adapt their parameters and architecture according to the difficulty of the input. While these networks show many benefits, such as the increased efficiency, the mitigation of overfitting \cite{overfitting}, overthinking \cite{shallowdeepnetworks, idk_cascades}, and vanishing gradient phenomena \cite{vanishinggradient}, many challenges are still open. In this paper, we specifically focus on a promising subset of this family, that is, Early Exit Neural Networks (EENNs) \cite{branchynet}. These architectures attempt to break the rigid \emph{one-size-fits-all} computation paradigm by allowing easy-to-classify samples to exit the network at intermediate layers via auxiliary classifiers, denoted as Early Exit Classifiers (EECs), bypassing the deeper blocks \cite{scardapane_why_2020, dynamax}. However, standard EENNs suffer from a fundamental reliability issue: the decision to exit is typically governed by a heuristic confidence threshold (e.g., softmax entropy) derived from a single model. Since DNNs are notoriously prone to overconfidence, even when making erroneous predictions, relying on a single agent self-assessment often leads to risky premature exits or unnecessary computation \cite{rethinking_calibration, guo2017calibration}. 

To address this shortcoming, we propose a shift to a collective confidence in early exits. Drawing inspiration from Ensemble Learning to reduce variance \cite{ensemblelearning}, we introduce SQUAD (Scalable Quorum Adaptive Decisions) that combines early exits with the ensemble technique. Instead of relying on the confidence of a single classifier, we query a dynamic committee of early-exit learners to decide whether to stop computation both horizontally (along the learners) and vertically (along the early exits). Indeed, in our framework, computation is halted horizontally, invoking each learner incrementally in order of computational complexity upon reaching a majority consensus, and vertically if the early exit decision is triggered once statistical robustness of the consensus is ensured through a t-test. This approach enhances the early exit decision with a statistically robust voting process, effectively mitigating the calibration issues of single-model EENNs. While conceptually appealing, implementing SQUAD is non-trivial. For a voting mechanism to be efficient, the committee members must offer complementary predictions not just at the final output, but at every intermediate early exit gate. We can define this property as \textit{hierarchical diversity}. This aspect is crucial since if all learners share the same biases or errors at layer $L$, the quorum will merely reinforce a wrong decision, and this is more likely to happen at the very early layers. 

Designing such a system manually is infeasible. Consequently, we could resort to Neural Architecture Search \cite{nasframework}, a method that is attracting steadily growing research interest for its ability to automatically design the architectures of NNs given a task and a dataset. Unlike traditional NAS, which optimizes for individual accuracy \cite{cnas, nachos, snas}, or ensemble methods that seek diversity only at the output \cite{nes, nesbs}, our approach explicitly searches for a set of learners that maximizes \textit{hierarchical diversity}. We term this unified framework QUEST (Quorum Ensemble Search Technique). By effectively combining the efficiency of early exits with the robustness of ensembles through a novel NAS strategy, QUEST offers a unique solution to the accuracy-efficiency dilemma. 

Lastly, empirical results on CIFAR-10 \cite{cifar}, CIFAR-100 \cite{cifar}, and ImageNet16-120 \cite{imagenet16} demonstrate that QUEST achieves a superior balance between accuracy and computational cost. Specifically, we improve inference time up to 70.60\% compared to static ensembles and increase accuracy up to 5.95\% over standard confidence-based EENNs at comparable latency. To facilitate comparisons and reproducibility, the source code of QUEST is released to the scientific community as a public repository\footnote{URL to be disclosed after paper acceptance}.

\section{Background}
\label{sec:background}

\label{sec:background}

\subsection{Early Exit Neural Networks}
Early Exit Neural Networks (EENNs), which allow intermediate predictions when enough confidence is achieved, are used to reduce the computational cost of deep neural networks \cite{scardapane_why_2020}.  An EENN can be viewed as a composition of $b$ sequential blocks, $f(x) = \mathrm{B}_{b} \circ \cdots \circ \mathrm{B}_{1}(x)$, where a classification head $C_i$ is attached to each block:
\begin{equation}
\label{eq:composition}
    f_i(x) = \mathrm{C}_{i} \circ \left( \mathrm{B}_{i} \circ \cdots \circ  \mathrm{B}_{1}(x) \right)
\end{equation}
This formulation produces a sequence of predictions, allowing the system to halt computation once a consensus or confidence threshold is met. Training is typically performed by jointly updating all exits:
\begin{equation}
    \label{eq:joint}
    L_{\text{joint}}(x, y) = l(f_b(x), y) + \sum_{i=1}^{b-1} \beta_i \cdot l(f_i(x), y)
\end{equation}
where $\beta_i$ weights the $i$-th branch loss and $x$ is the data input. 

A weakness of this technique is that confidence thresholds are often unreliable due to calibration issues \cite{rethinking_calibration, guo2017calibration}. A metric of calibration is the Expected Calibration Error (ECE) \cite{naeini2015obtaining}, which measures the reliability of a model by quantifying the expected difference between its predicted confidence and its actual accuracy. To calculate this, the full probability range is divided into $M$ disjoint bins (i.e., intervals of size $1/M$), and samples are grouped into these bins based on their predicted confidence scores. For each bin $B_m$, the average accuracy $\text{acc}(B_m)$ and average confidence $\text{conf}(B_m)$ are computed. The final ECE score is the weighted average of the absolute difference between these two values, weighted by the fraction of samples $\frac{|B_m|}{N}$ in each bin, as shown in the formula:

\begin{equation}
\text{ECE} = \sum_{m=1}^{M} \frac{|B_m|}{N} \left| \text{acc}(B_m) - \text{conf}(B_m) \right|
\end{equation}

\subsection{Ensemble Learning and Consensus}
A way to improve the calibration of deep neural networks is by Ensemble Learning \cite{ensemblelearning}, 
where multiple models (learners) are trained to solve the same task and their predictions are aggregated, averaging out individual errors \cite{averagingerror}. 

 To leverage this diversity, a consensus mechanism is required. 
Soft voting,  which averages the predicted probability distributions, is generally preferred for DNNs as it weighs learner confidence, penalizing uncertainty more effectively than simple hard voting. 
However, standard voting introduces a rigid computational bottleneck, as inference cost scales linearly with ensemble size $M$. To quantify the diversity necessary for effective ensembles, we utilize the Pairwise Predictive Disagreement (PPD) \cite{ppd}. Let $\mathcal{E} = \{f_1, \dots, f_M\}$ be an ensemble and $\mathcal{D}$ a dataset of $N$ samples. The PPD is defined as:
\begin{equation}
    \text{PPD}(\mathcal{E}, \mathcal{D}) = \frac{1}{\binom{M}{2}} \sum_{1 \le i < j \le M} \left( \frac{1}{N} \sum_{k=1}^{N} \mathbb{I}\left(\hat{y}_i^{(k)} \neq \hat{y}_j^{(k)}\right) \right)
\end{equation}
where a higher PPD value indicates greater diversity in predictive behavior, essential for robust consensus.

\subsection{DARTS, the Neural Architecture Search Space}
\label{subsec:darts}
DARTS ~\cite{liu_darts_2019} is a cell-based NAS framework, where each cell is represented as a directed acyclic graph (DAG) with~$N$ nodes arranged sequentially and edges connecting them, where operations on each edge are drawn from a shared set of possible operations. 
Let $\mathcal{O}=\{o^{(i,j)}\}$ be the set of candidate operations between the $i$-th and $j$-th nodes. DARTS assigns continuous architecture parameters $\balpha=\{\alpha^{(i,j)}\}$, where each $\alpha^{(i,j)} \in \mathbb{R}^{|\mathcal{O}|}$ weights the choice of operation on a given edge in a weighted sum of operations, and must be optimized. 
DARTS space consists of two types of cells: \emph{normal cells}, which preserve spatial dimensions (height and width), and \emph{reduction cells}, which reduce them and increase feature depth. 
After searching, the final, discrete architecture is constructed by retaining the strongest~$k$ operations (from distinct nodes) among all non-zero candidates for each intermediate node. 
The total number
of architectures is approximately $10^{18}$.

\subsection{Neural Architecture Search via Bayesian Sampling}
Constructing an ensemble with high PPD and accuracy is a combinatorial problem with NAS search spaces being exponentially large. Neural Ensemble Search via Bayesian Sampling (NESBS) \cite{nesbs} addresses this by finding $n$ optimal candidate NNs efficiently.  NESBS employs a three-phase process: it uses a supernet to estimate performance via SPOS~\cite{spos}, derives a posterior distribution of architectures, and finally exploits Stein Variational Gradient Descent with Regularized Diversity (SVGD-RD) to sample ensembles.
The core mechanism, SVGD-RD, updates architectural configurations using an update direction $\phi^*$ composed of two opposing forces.
The \textit{Driving Force} steers particles toward high-accuracy regions (exploiting the supernet learned distribution), while the \textit{Repulsive Force} creates a dispersive vector field to push particles apart, ensuring diversity. The coefficient $\delta$ amplifies this repulsion to prevent mode collapse.

\section{Related literature}
\label{sec:relatedliterature}

\subsection{Early exit methodologies}

BranchyNet \cite{branchynet} introduces EECs via joint training, a paradigm later extended by recursive strategies \cite{scardapane_differentiable_2020} and distillation-based approaches like QUTE \cite{QUTE}. While other works interpret early exits as implicit, weight-sharing ensembles \cite{earlyexitensemble,QUTE, depthuncertainty, MIMMO}, they typically aggregate intermediate predictions to refine uncertainty or accuracy. Crucially, unlike SQUAD, these approaches leverage the ensemble for performance gains, neglecting the dynamic computation halting used to reduce inference time.

\subsection{MACs-Aware NAS}
The first group designs static neural networks. Prominent solutions are CNAS~\cite{cnas} that optimizes architectures by explicitly imposing MACs constraint and SNAS~\cite{snas} that maximizes expected accuracy over a distribution of graphs and applies a penalty to reduce the MACs, respectively. 
The second group designs Dynamic Neural Networks, particularly EENNs, which introduce runtime adaptability. A state-of-the-art example is NACHOS~\cite{nachos}, which searches for Pareto-optimal EENNs by jointly optimizing the backbone, exit placements, and thresholds under MAC constraints and exploits the differentiable training mentioned above \cite{scardapane_differentiable_2020}. Other approaches simplify the search scope by using frozen backbones (NASEREX~\cite{naserex}) or employing genetic algorithms for multi-objective optimization without constraints (EDANAS~\cite{gambella2023edanas}).

\subsection{NAS for Ensembles} 

These NAS solutions shift focus from single models to committees. 
More recently, NES~\cite{nes} introduced diversity-aware selection, though with high search costs. Subsequent works addressed efficiency: NESBS~\cite{nesbs} significantly reduced search cost via Bayesian sampling, NEAS~\cite{neas} enabled layer sharing, and SAEP~\cite{saep} integrated pruning. However, these methods optimize diversity primarily at the output level, overlooking the \textit{hierarchical} diversity required for robust early-exit consensus.

\section{The proposed inference of SQUAD}
\label{sec:quorum}

\subsection{The architecture}

\begin{figure}[t]
    
    \centering
    \scalebox{0.4}{
      
        \includegraphics[width=\textwidth]{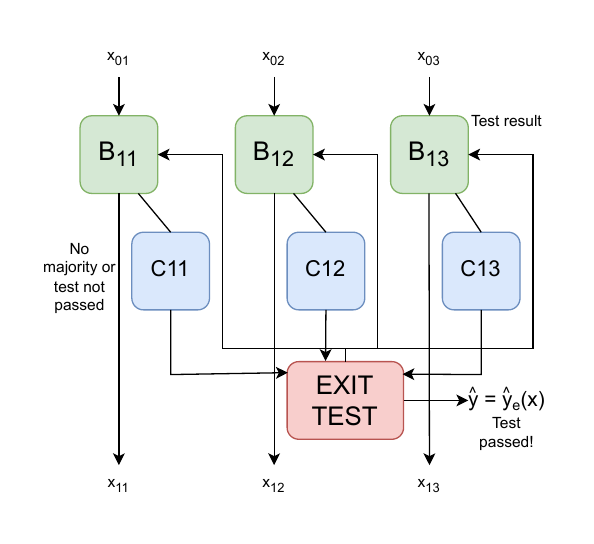}
    }
    \caption{Overview of the first stage ($e$=1) of SQUAD with $K=3$ and $k\in\{1,...,3\}$ denotes the branch index. The Exit Test incrementally collects the class confidences from the EECs denoted as $C_{1k}$, ordered by the computational cost of $B_{1k}$. If the stopping criterion of the Exit Test is satisfied, an intermediate prediction is done, and the input representations are not forwarded to the following branches. Otherwise, the output of the blocks $B_{1k}$ is propagated through the following blocks. The test result acts as a stopping signal when a decision in the Exit Test has been taken.}
    \label{fig:quorum_scheme}
\end{figure}

The architecture of SQUAD consists of an ensemble of $K$ parallel learners, denoted as $\{L_1, \dots, L_K\}$, each equipped with $E$ intermediate exit gates indexed by $e \in \{1, \dots, E\}$. At each exit stage $e$, we define a set of \textit{branches} $B_e = \{b_{e1}, \dots, b_{eK}\}$, where each $b_{ek}$ represents the block of backbone layers in learner $k$ leading to its respective EEC, denoted as $C_{ek}$. Notably, the learners operate with independent weights (no parameter sharing). The overall inference process is designed such that the input sample $x$ is processed in parallel across the distinct backbones, while the exit decisions are evaluated sequentially from $e = 1$ to $E$. The schematic of a single stage of this framework is illustrated in Figure \ref{fig:quorum_scheme}.

\subsection{Inference logic}
\label{subsec:inferencesquad}
At each stage $e$, the intermediate predictions are aggregated using a MACs-aware protocol. This approach is grounded in the rationale that, although samples are forwarded in parallel, learners exhibit varying latencies. To simulate this \textit{first-come first-vote} mechanism, we estimate latency via Multiply-Accumulate operations (MACs). Learners contribute to the voting pool incrementally, ordered by increasing computational complexity, until a majority consensus is formed. The inference terminates at exit $e$ only if this consensus satisfies the specific statistical criteria defined by the t-test validation.
\subsection{Majority voting}
The voting process is incremental, with the previously defined order. Let $f_{e,k}(x)$ denote the output probability distribution (softmax) of the $k$-th branch at exit $e$. Each vote, specifying the predicted class, is:
\begin{equation}
    \hat{y}_{e,k} = \arg\max_{c} f_{e,k}(x)_c
\end{equation}
Let $V_e^{(m)} \in \mathbb{R}^C$ be the vote vector after $m$ branches have voted:
\begin{equation}
    V_e^{(m)} = \sum_{j=1}^{m} \text{OneHot}(\hat{y}_{e, j})
\end{equation}
A \textit{pivot branch} $b_{pivot}$ is identified as the first branch $m$ where the voting outcome is decided irrespectively by the remaining $K-m$ branches. The pivot is found when: 
\begin{enumerate}
    \item \textbf{Quorum reached}: The leading class has $\ge \lfloor K/2 \rfloor + 1$ votes.
    \item \textbf{Quorum unfeasibility}: with the remaining votes, no class can satisfy the above condition and so cannot meet the quorum.
\end{enumerate}
The consensus class at exit $e$ is defined as $\hat{y}_e = \arg\max V_e^{(pivot)}$. Once the pivot is found, if \textbf{Quorum reached} applies, the network proceed to evaluate early exit condition, while if \textbf{Quorum unfeasibility} happens, the input representations are forwarded directly to the next K branches since there is no agreement, hence saving useless computation.


\subsection{Early Exit Condition (t-test)}
\label{subsec:tstudent}
To validate the significance of the votes for the winning class (if any), we resort to a t-test as our early exit trigger. The majority voting acts as a stronger requirement for the winning class than the soft voting, but could hide the uncertainty of the learners, which is screened by the t-test.

Once the consensus class $\hat{y}_e$ is determined, we evaluate the confidence of the prediction to decide whether to exit (positive) or to forward the input representations to the next K branches (negative). We consider the subset of branches that have voted for the consensus class, denoted as $S_e = \{k \in \{1,\dots,K\} \mid \hat{y}_{e,k} = \hat{y}_e\}$. The confidence scores for this subset are $P_{S_e} = \{ \max f_{e,k}(x) \mid k \in S_e \}$.

The simplest approach to trigger early exit would be to check that the mean of these confidence scores exceeds a threshold $\tau_{\text{conf}}$. However, to ensure statistical robustness, we compute the Lower Confidence Bound (LCB) using the t-test as follows
\begin{equation}
    \text{LCB}_e = \mu_{S_e} - t_{\alpha, |S_e|-1} \cdot \frac{\sigma_{S_e}}{\sqrt{|S_e|}}
    \label{eq:lcb}
\end{equation}
where $\mu_{S_e}$ is the sample mean, $|S_e|$ is the sample size, and $t_{\alpha, |S_e|-1}$ is the critical t-value for a one-sided 95\% confidence interval with $|S_e|-1$ degrees of freedom. Crucially, $\sigma_{S_e}$ denotes the \textit{unbiased sample standard deviation} (calculated using Bessel's correction with $|S_e|-1$ in the denominator) to ensure statistical robustness.

The inference terminates at exit $e$ if:
\begin{equation}
    \text{LCB}_e > \tau_{\text{conf}}
\end{equation}
where $\tau_{\text{conf}}$ is a user-defined confidence threshold. If the condition is met, the final prediction is $\hat{y}_e$. Otherwise, the input representations proceed to exit $e+1$. If the last exit $E$ is reached without triggering, the prediction at exit $E$ is forced.

\section{The proposed QUEST}
\label{sec:quest}

\subsection{Problem formulation}
\label{sct:problem_formulation}

Designing SQUAD manually is infeasible, therefore we introduce in this section the QUEST NAS strategy to automate quorum-based EENN implementation.

The problem addressed by QUEST consists of selecting the set of EENNs aiming at providing a high accuracy with their adaptive ensemble inference, by optimizing their individual accuracy and representational diversity across the early exit gates. Formally, QUEST can be defined as the following joint optimization problem: 
\begin{align}
\label{eq:problem}
    \text{minimize }  &\mathcal{G} \left 
    ( \mathcal{F_A}(\{\tilde{x}_1, .., \tilde{x}_E\}),  
   \mathcal{F_D}(\{\tilde{x}_1, .., \tilde{x}_E\}) 
    \right )
    \\
    \text{s. t. }  & \tilde{x} \in \Omega_{\tilde{x}}
    \nonumber
\end{align}
where $\mathcal{G}$ is a bi-objective optimization function, $\tilde{x}$ represents a candidate EENN architecture, $\Omega_{\tilde{x}}$ is the supernet enhanced with early exits, $\mathcal{F_A}(\{\tilde{x}_1, .., \tilde{x}_E\} )$ is the classification accuracy of the ensemble and $\mathcal{F_D}(\{\tilde{x}_1, .., \tilde{x}_E\})$ is the representational diversity of the ensemble members. 

\subsection{Overview of QUEST}

\begin{figure*}[t]
    \centering
    \scalebox{0.8}{
    \includegraphics[width=\textwidth]{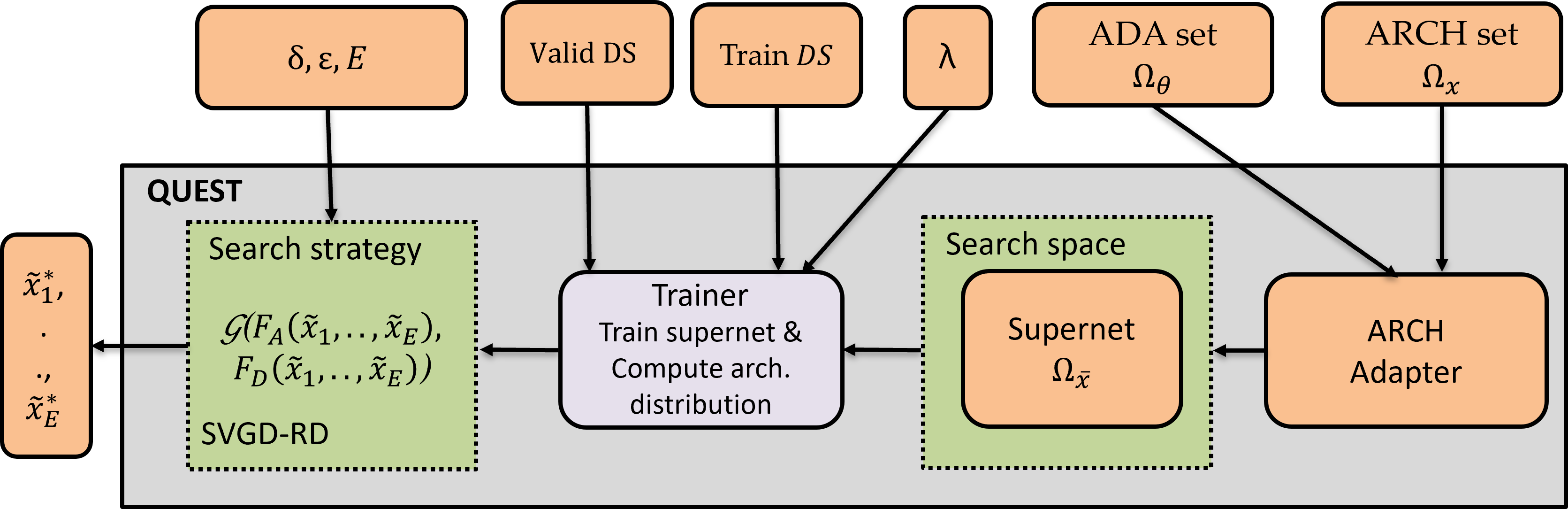}
    }
    \caption{Overview of QUEST consisting of the Arch Adapter, the Search Space, the Trainer, and the Search Strategy.}
    \label{fig:QUEST_scheme}
\end{figure*}

An overall description of the QUEST framework is given in Fig.~\ref{fig:QUEST_scheme}. QUEST receives as input a dataset $\mathcal{DS}$, comprising a training part (used to train the candidate networks) and a validation part (used to validate the candidate networks), a supernet $\Omega_x$ from which the candidate neural networks are obtained, and the ADA set $\Omega_{\theta}$ from which the EENN parameters are obtained. 

At the end of the search, QUEST returns the set of the $k$ optimal EENN architectures $\tilde{X}^* = \{ \tilde{x}_1^*, \ldots, \tilde{x}_E^* \}$ for a dynamic ensemble, being $E$ a value specified by the user. We emphasize that the output of QUEST is a set of networks enhanced with EECs, while the supernet given in input consists of single-exit neural networks (i.e., no EECs).

The Arch Adapter receives a supernet $\Omega_x$ and an ADA set $\Omega_\theta$ specifying the number and the placement of the early exits along the backbone of the supernet. The module outputs the resulting supernet $\Omega_{\tilde{x}}$ enriched with early exit classifiers placed.

We follow the three-phase process of NESBS, as introduced in Section \ref{sec:background}, where the Trainer accounts for the first two phases. 

\subsection{Training of the supernet with early exits}

The Trainer receives the supernet $\Omega_{\tilde{x}}$ and the dataset $\mathcal{DS}$. The dataset $\mathcal{DS}$ is divided into a \textit{Train DS}, used in the first phase, and a \textit{Valid DS}, used in the second phase.
It outputs a set $\alpha$, being matrices describing the variational posterior distribution of the architectural choices of the network, with each matrix referring to a particular cell of the searched network as in the definition of DARTS search space \ref{subsec:darts}.

The first step is to train the early exit supernet $\Omega_{\tilde{x}}$ on the \textit{Train DS} by using an early exit variation of SNAS, which will be introduced later in this section, to find an architectural distribution $\alpha$ that favors operations that bring high accuracy and moderate complexity (in terms of MACs) in the searched network. As SNAS, the training optimizes the weights $w$ and the architectural parameters $\alpha$ of the supernet on the objective function computed as the expected cost of the sum an accuracy term and a penalty term over a distribution of architectures $\alpha$:
\begin{equation}
\label{eq1}
\mathbb{E}_{\tilde{x}_p \sim p(\alpha)}[L(w^{*}(\alpha),\alpha)=L_{joint}(w^{*}(\alpha),\alpha) + \lambda \times L_C(\alpha)]
\end{equation}
with $\tilde{x}_p$ being a sampled architecture (i.e., a one-hot vector per edge).

Different from SNAS, we compute the accuracy term differently from SNAS since $\tilde{x}_p$ is an early exit architecture. Indeed, $L_{joint}(w^{*}(\alpha),\alpha)$ is the sum of the validation loss of each exit of $\tilde{x}_p$ by following the Joint Training scheme as in Eq. \ref{eq:joint}, with each validation loss $L^{i}(w^{*}(\alpha),\alpha)$ computed with the $i^{th}$ EEC over the sampled backbone $x_p$. This adapts SNAS to a supernet with multiple exits.

The penalty term $L_C(\alpha)$ follows the SNAS definition and computes a penalty according to the expected cost in terms of MACs over the distribution of $\alpha$. $L_C(\alpha)$ is the sum of the terms $L^{n}_C(\boldsymbol{\alpha})$, defined at every $n^{th}$ searched cell of the searched network as follows:
\begin{equation}
\label{penaltycost}
L^{n}_C(\boldsymbol{\alpha}) = \sum_{i} \sum_{j}\sum_{k} \text{MACs}^{(i,j)}_k \cdot x^{(i,j)}_k(\alpha)
\end{equation}
where $x^{i,j}_k$ is the softened one-hot random variable for operation selection at edge $(i,j)$ with k being the operation index, and $\text{MACs}^{(i,j)}_k$ the corresponding number of MACs. 

In this preliminary step, we use SNAS instead of SPOS, which uses uniform sampling and is adopted in NESBS, for its ability to find good architectures with moderate computational complexity, a crucial property of an efficient neural network, and model our architectural distribution.

\subsection{Computation of the variational posterior distribution}

The computation of the variational posterior distribution is explained in Algorithm \ref{alg:variational_search}. Differently from NESBS, we applied the Joint Training Loss, explained in \eqref{eq:joint}, to compute the negative likelihood (NLL). By backpropagating gradients from all exits simultaneously, the search steers the variational distribution toward operations that maximize discriminative power at varying depths. The Kullback-Leibler divergence term ensures high entropy in the architectural parameters to avoid collapse to specific operations.

\begin{algorithm}[h!]
\caption{Variational Architecture Search for QUEST}
\label{alg:variational_search}
\begin{algorithmic}[1]
\REQUIRE Validation dataset \textit{Valid DS}, Prior distribution $p(\boldsymbol{\alpha})$ (Uniform), KL weight $\beta$, Learning rate $\eta$
\ENSURE Optimized variational parameters $\phi^*$ defining the posterior $q_{\phi}(\boldsymbol{\alpha})$

\STATE \textbf{Initialize:} Variational parameters $\phi$ (e.g., logits of operations)

\WHILE{not converged}
    \STATE \textit{\% 1. Reparameterization Trick}
    \STATE Sample noise $\boldsymbol{\epsilon} \sim p(\boldsymbol{\epsilon})$ \quad (Gumbel)
    \STATE Generate architecture sample $\boldsymbol{\alpha} = g(\phi, \boldsymbol{\epsilon})$ differentiable w.r.t $\phi$
    
    \STATE \textit{\% 2. Compute Vertical Efficiency Loss}
    \STATE Compute Joint Loss on Validation Set (Hierarchical Constraint):
    \STATE \quad $\mathcal{L}_{NLL} \leftarrow L_{joint}(\boldsymbol{\alpha}, \textit{Valid DS})$ \quad \textit{(via Eq. \ref{eq:joint})}
    
    \STATE \textit{\% 3. Compute Regularization}
    \STATE Calculate KL Divergence against Uniform Prior:
    \STATE \quad $\mathcal{L}_{KL} \leftarrow D_{KL}(q_{\phi}(\boldsymbol{\alpha}) || p(\boldsymbol{\alpha}))$
    
    \STATE \textit{\% 4. Objective Construction}
    \STATE Combine terms for Negative ELBO:
    \STATE \quad $\mathcal{L}_{ELBO} \leftarrow \mathcal{L}_{NLL} + \beta \mathcal{L}_{KL}$
    
    \STATE \textit{\% 5. Optimization Step}
    \STATE Update variational parameters via gradient descent:
    \STATE \quad $\phi \leftarrow \phi - \eta \nabla_{\phi} \mathcal{L}_{ELBO}$
\ENDWHILE

\textbf{return} $\phi$
\end{algorithmic}
\end{algorithm}

\subsection{Search Strategy via SVGD-RD}

As shown in NESBS, we can now use Stein Variational Gradient Descent with regularized diversity (SVGD-RD), to effectively and efficiently optimize the accuracy $\mathcal{F}_A$ (the driving force) and the diversitiy $\mathcal{F}_D$ (the repulsion force), as introduced in Eq. \ref{sec:background}.
However, different from NESBS, our framework leads NESBS to find an ensemble of early exit neural networks thanks to the way we modeled the variational posterior distribution.

\begin{table*}[ht!]
    \centering
    \caption{Performance on CIFAR-10, CIFAR-100, and ImageNet16-120.}
    \label{tab:dataset_comparison}
    \resizebox{\textwidth}{!}{
    \begin{tabular}{lccccccccccccc}
        \toprule
        & \multicolumn{3}{c}{\textbf{CIFAR-10}} & & \multicolumn{3}{c}{\textbf{CIFAR-100}} & & \multicolumn{3}{c}{\textbf{ImageNet16-120}} & \textbf{ Search Time (h)} \\
        \cmidrule{2-4} \cmidrule{6-8} \cmidrule{10-12}
        \textbf{Method} & \textbf{Acc (\%)} & $\boldsymbol{F_M (M)}$ & $\boldsymbol{F_{MT} (M)}$ & & \textbf{Acc (\%)} & $\boldsymbol{F_M (M)}$ & $\boldsymbol{F_{MT} (M)}$ & & \textbf{Acc (\%)} & $\boldsymbol{F_M (M)}$ & $\boldsymbol{F_{MT} (M)}$ & & \\
        \midrule
        NESBS  & 97.64 & 694.02 & 2013.94 & & 85.45 & 694.08 & 2014.12 & & 56.15 & 161.09 & 503.69 &  8 \\

        NACHOS & 95.40 & 212.95 & 212.95 & & 79.25 & 312.14 & 312.14 & & 51.10 & 95.53 & 95.53 & 36h \\
        CNAS   & 93.73 & 216.04 & 216.04 & & 80.23 & 315.32 & 315.32 & & 51.38 & 97.60 & 97.60 &  24h \\
        \midrule 
        $\text{QUEST}_\text{Test}$  & 96.97 & 205.87 & 603.63 & & \textbf{82.20} & \textbf{302.87} & 878.73 & & \textbf{57.05} & \textbf{93.12} & 267.71 & 8 \\
        $\text{QUEST}_\text{Mean}$ &  \textbf{97.03} & \textbf{204.06} & 598.23 & & 79.97 & 226.45 & 656.72 & & 54.22 & 61.75 & 177.37 & 8h \\
        \bottomrule
    \end{tabular}
    }
\end{table*}

\section{Experimental campaign}
\label{sec:results}

\subsection{Computational metrics}

MACs have been computed to define two computational metrics. The metric $F_M$ refers to the number of MACs after we achieve a response, and in an ensemble, is equal to the sum of the maximum number of MACs among the voting members of the ensemble at each exit. It is a metric related to the latency. In the following formulations, $K$ denotes the total number of parallel branches in the ensemble, while $E^*$ represents the specific exit where the sample decision is finalized. The term $C_{k,e}$ defines the incremental MAC cost associated with branch $k$ at exit $e$. We employ $\pi$ as a permutation that sorts the branches in non-descending order of costs ($C_{\pi(0),e} \leq \dots \leq C_{\pi(K-1),e}$), to follow the \textit{first-come first-vote} logic introduced in \ref{subsec:inferencesquad}, allowing us to identify $m$, the rank of the \textit{pivot branch} that triggered the exit decision.
\begin{equation}
    F_M = \sum_{e=0}^{E^*}\left(\max_{k \in \{0, \dots, m\}} \{ C_{\pi(k),e} \} = C_{\pi(m),e}\right)
\end{equation}
The metric $F_{MT}$ refers to the effective total number of MACs used to achieve a response, and in an ensemble, is equal to the sum of the total amount of MACs used by the members of the ensemble up to the final vote at each exit. It is a metric related to energy consumption. 
\begin{equation}
    F_{MT} = \sum_{e=0}^{E^*}\sum_{k=0}^{m} C_{\pi(k),e} + (K - 1 - m) \cdot C_{\pi(m),e}
\end{equation}
Consequently, the energy metric comprises two components: the \textit{completed work} ($\sum_{k=0}^{m} C_{\pi(k),e}$), which sums the costs of the pivot branch and all computationally lighter branches that finished execution; and the \textit{parallel overhead} ($(K - 1 - m) \cdot C_{\pi(m),e}$), which accounts for the computational work leaked by the heavier background branches that were interrupted exactly when the pivot branch finished.

\subsection{DARTS setting}
\label{subsec:setting}
For experiments conducted in the DARTS \cite{liu_darts_2019} search space, we follow the original search settings and the evaluation settings of SNAS and NESBS. The hyperparameters of the search and the training are shown in Appendix \ref{sec:hyperparameters}.
Moreover, the architecture search is performed on CIFAR-10, and the resulting genotype is then evaluated on various datasets by training with early exits the corresponding neural networks from scratch, using an early exit weight $\beta_i=1$, introduced in \ref{eq:joint}, for all the EECs, both during the training and search phase. We highlight that during the search phase, the explored networks are smaller in terms of size with respect to the final trained network to balance efficiency and performance of the NAS. This is shown by the different number of channels and cells stacked in Table \ref{tab:settingsupernetdarts} and Table \ref{tab:settingtraindarts} in Appendix \ref{sec:hyperparameters}.

\subsection{Experimental results}

For our experiments we use the following hyperparameters. 
The number of ensemble members $E=3$. The ARCH set defines the search space of DARTS, resulting in each branch $B_{ek}$ of our framework being a group of normal cells and a reduction cell. The ADA set specifies that the EECs are placed after a reduction cell. The architecture of each EEC is a stack of Global Average Pooling, Flatten, and a Linear layer. The datasets used in our experimental campaign are CIFAR-10, CIFAR-100, and ImageNet16-120, standard NAS benchmarks with increasing difficulty explained in Appendix \ref{sec:datasets}. For our QUEST models, we selected a confidence threshold $\tau_{conf}\in\{0.95, 0.6, 0.3\}$ through grid search in our three datasets, respectively, by selecting a reasonable threshold of desired accuracy.



We run the search of QUEST on CIFAR-10 and we then compute CNAS and NACHOS with a constraint on the number of MACs equal to the $F_M$ of the found ensembles by QUEST, to have a static model and a dynamic model for comparison. We also added NESBS for a static ensemble as a baseline. The search spaces explored by NESBS and QUEST is DARTS, while CNAS and NACHOS look for OFA-based MobileNetV3 models, a state-of-the-art family in edge solutions.  Different from QUEST and NESBS being run once on CIFAR-10, as explained in Section \ref{subsec:setting}, CNAS and NACHOS are run for each dataset since we use a different MACs constraint. Lastly, we reported the performance of the set of EENNs found by QUEST tested by using the SQUAD framework with two different exit criteria, mean confidence and t-test, explained in Sec. \ref{subsec:tstudent}, as $\text{QUEST}_\text{Mean}$ and $\text{QUEST}_\text{Test}$, respectively.



Our main results are introduced in  Table \ref{tab:dataset_comparison}, where \textit{Search Time} refers to the GPU hours used by the search of the NAS methods to find the examined models. 
We show that, despite a drop in accuracy, QUEST is able to significantly reduce both $F_M$ and $F_{MT}$ with respect to the static ensemble found by NESBS. In particular, $F_M$ is reduced by 70.60\%, 56.36\%, 42.19\% on CIFAR-10, CIFAR-100, ImageNet16-120, respectively.
Single models found by CNAS and NACHOS show lower accuracy with comparable $F_M$ with respect to QUEST, while have better $F_{MT}$ since they require a forward pass for a single model, different from ensembles as shown in \ref{sec:background}. The accuracy gain of QUEST with respect to NACHOS is by 1.63\%, 2.95\%, 5.95\% on CIFAR-10, CIFAR-100, and ImageNet16-120, respectively.

\begin{figure*}[h!]
    \centering

        \includegraphics[width=\textwidth]{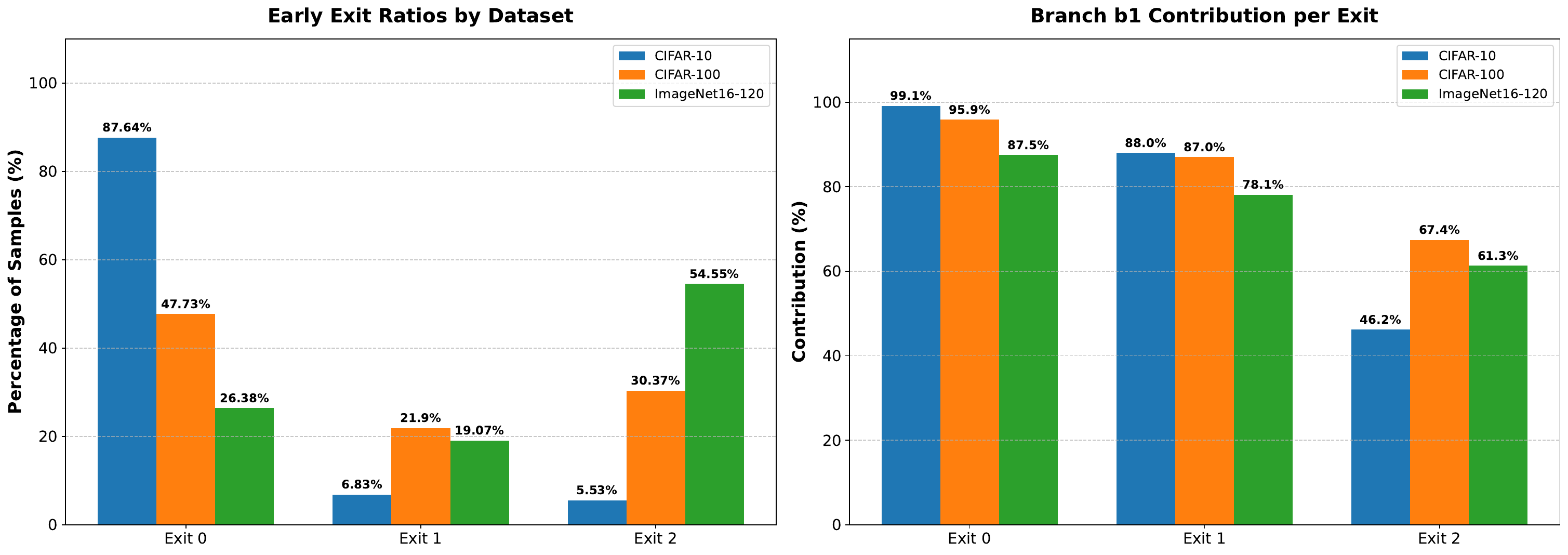}
    \caption{Left panel: Early exit ratios by dataset. The majority of samples in simple datasets (CIFAR-10) exit early, while complex datasets (ImageNet) require deeper processing. Right panel: Branch pivot (the decisive vote) ratios by dataset. Branch b1 dominates for easy samples, while Branch b2 becomes critical for reaching consensus on hard samples.}
    \label{fig:earlyexitratios}
\end{figure*}

We analyze the vertical dynamism of our framework in
Figure \ref{fig:earlyexitratios}. A clear trend emerges: the distribution of exit events shifts deeper into the network as the classification task becomes more challenging. 
Additionally, we analyze the horizontal dynamism of our framework in
Figure \ref{fig:earlyexitratios}. We refer to the branches of our ensemble as b0, b1, b2. We plot only the usages of b1 because b0 cannot be the pivot due to the requirement of at least two votes (quorum), and b2 usages are just complementary to b1. As the most efficient learner in the ensemble (sorted by MACs), b1 acts as the primary driver of the consensus mechanism, consistently accounting for the majority of the voting weight. However, a clear inverse correlation is observable between b1 dominance and the complexity of the inference task. As a result, b2 is effectively recruited as a consultant only for harder samples. 
To sum up, the two dynamism effectively adapt to the complexity of the classification task.

Additionally, we quantitatively estimated and show in Table~\ref{tab:ppd_compact} the novel concept of hierarchical diversity by computing the average PPD, the metric for diversity in an ensemble introduced in Section~\ref{sec:background} over all the exits of our QUEST ensemble. The results show that the QUEST ensemble, at every exit, has a better PPD than NESBS, validating the statement that QUEST is able to find ensembles with a good hierarchical diversity.

\begin{table}[h]
\centering
\caption{Comparison of PPD (\%) across datasets. Note: NESBS only utilizes the final exit.}
\label{tab:ppd_compact}
\scalebox{0.85}{
\begin{tabular}{lcccc}
\toprule
\textbf{Dataset} & \textbf{Exit} & \textbf{QUEST} & \textbf{NESBS} \\
\midrule
\multirow{3}{*}{CIFAR-10} 
 & Exit 0 & 4.08 & -- \\
 & Exit 1 & 3.13 & -- \\
 & Exit 2 & \textbf{3.15} & 2.27 \\
\midrule
\multirow{3}{*}{CIFAR-100} 
 & Exit 0 & 21.30 & -- \\
 & Exit 1 & 17.25 & -- \\
 & Exit 2 & \textbf{17.50} & 13.16 \\
\midrule
\multirow{3}{*}{ImageNet16} 
 & Exit 0 & 37.91 & -- \\
 & Exit 1 & 37.34 & -- \\
 & Exit 2 & \textbf{39.18} & 35.22 \\
\bottomrule
\end{tabular}
}
\end{table}

\begin{table}[h]
\centering
\caption{The first group represents the per-exit ECE (\%) of QUEST across datasets. The second group represents the overall ECE (\%) of different models across datasets. }
\label{tab:ece_results}
\scalebox{0.85}{
\begin{tabular}{lccc}
\toprule
\textbf{Model} & \textbf{CIFAR-10} & \textbf{CIFAR-100} & \textbf{ImageNet16} \\
\midrule
QUEST (0)  & 0.60 & 0.76 & 4.89 \\
QUEST (1)  & 6.37 & 6.53 & 4.38 \\
QUEST (2)  & 13.74 & 21.50 & 13.23 \\
\midrule
QUEST (all) & \textbf{1.72} & \textbf{8.30} & \textbf{6.72} \\
NACHOS & 5.33  & 9.15 & 7.54 \\
QUTE & 2.50 & - & - \\
EE-ensemble & 3.30 & - & - \\
\bottomrule
\end{tabular}
}
\end{table}

Also, we measured the Expected Calibration Error of our QUEST model. The table \ref{tab:ece_results} early exits are well-calibrated. Indeed, for CIFAR-10 and CIFAR-100, the first exit (Exit 0) has extremely low ECE ($<1.00$). This confirms that when the model is confident enough to exit early, it is almost perfectly reliable. Across all datasets, the final exit has the highest calibration error (e.g., 21.5\% on CIFAR-100). This confirms the hypothesis that the samples at the last exit result in overconfident predictions. 
Overall, SQUAD shows better calibration than QUTE \cite{QUTE} and EE-ensemble \cite{earlyexitensemble} on CIFAR-10 and on NACHOS on all three datasets.

We show the performance of our QUEST model under varying $\tau_{conf}$ and exit criterion, being mean confidence or the t-test on the three different datasets in Appendix \ref{sec:exitcriteria} respectively. In general, as expected, the t-test is more strict in triggering early exiting, while the mean confidence is more permissive, resulting in the latter solutions having lower accuracy and $F_M$. Interestingly, on CIFAR-10, $\tau_{conf}=0.95$ shows a slightly better trade-off than the other exit criterion, suggesting that a simple exit method could be sufficient for an easier (considering the earlier exits) task.

\section{Conclusions} 
\label{sec:conclusions}
In this work, we introduced SQUAD, a framework that reconciles the accuracy of ensemble learning with the efficiency of dynamic neural networks. By coupling a sequential Quorum-based voting mechanism with Early Exit backbones, SQUAD is able to adaptively use only a subset of learners invoked incrementally in order of computational cost, but also a fraction of their backbones, exploiting the early exits. We also define a NAS to find a good ensemble of EENNs for SQUAD by optimizing the accuracy and the hierarchical diversity jointly, ensuring learners are complementary at every depth. Experiments on CIFAR-10, CIFAR-100, and ImageNet16-120 demonstrate that QUEST achieves state-of-the-art trade-offs, significantly reducing MACs in parallel inference compared to static ensembles while outperforming single-model dynamic solutions. 

\newpage
\bibliography{icml}
\bibliographystyle{icml2026}

\newpage
\appendix
\onecolumn

\section{Differentiable Stochastic Neural Architecture Search}

Stochastic NAS (SNAS) \cite{snas} assigns continuous architecture parameters $\boldsymbol{\alpha}^{(i,j)}$, which parametrize the distribution of operation selection.
To maintain differentiability while mimicking discrete sampling, the \textit{mixed operation} is defined as the stochastic weighted sum:
\begin{equation}
    \label{eq:snas_mixedop}
    \bar{o}^{(i,j)}(x) \;=\; \sum_{k=1}^{|\mathcal{O}|} Z_{k}^{(i,j)} \cdot o_k(x),
\end{equation}
where $Z_{k}^{(i,j)}$ is a variable sampled from the \textit{Concrete distribution} via the reparameterization trick:
\begin{equation}
    Z_{k}^{(i,j)} = \frac{\exp\left(\left(\log \alpha_{k}^{(i,j)} + G_{k}^{(i,j)}\right)/\lambda\right)}{\sum_{l=1}^{|\mathcal{O}|} \exp\left(\left(\log \alpha_{l}^{(i,j)} + G_{l}^{(i,j)}\right)/\lambda\right)}.
\end{equation}
Here, $G_{k}^{(i,j)} = -\log(-\log(U_{k}^{(i,j)}))$ represents Gumbel noise drawn from a uniform distribution $U \sim \text{Uniform}(0,1)$, and $\lambda$ is the softmax temperature.
Unlike DARTS, which requires a heuristic \emph{argmax} selection post-search, SNAS steadily anneals $\lambda \to 0$ during training. This forces the random variable $\boldsymbol{Z}^{(i,j)}$ to converge to a one-hot vector, making the mixed operation unbiased and effectively discrete by the end of optimization.

\section{NESBS}

The core of NESBS \cite{nesbs} is SVGD-RD, which basically updates $n$ architectural configurations using SGD learning through the update direction $\phi^*$, which drives the evolution of the architecture particles and is composed of two distinct mechanisms that balance individual performance with collective diversity.

\begin{equation}
\phi^*(\tilde{x}) = \frac{1}{N} \sum_{j=1}^{N} \bigg[  \underbrace{k(\tilde{x}_j, \tilde{x}) \nabla_{\tilde{x}_j} \log p(\tilde{x}_j)}_{\text{Driving Force}}
 + \underbrace{(1 + \delta) \nabla_{\tilde{x}_j} k(\tilde{x}_j, \tilde{x})}_{\text{Repulsive Force}} \bigg]
\end{equation}

\begin{itemize}
    \item \textbf{Driving Force (Accuracy $\mathcal{F}_A)$}: The first term, $k(\tilde{x}_j, \tilde{x}) \nabla_{\tilde{x}_j} \log p(\tilde{x}_j)$, acts as the exploitative component of the optimization. It leverages the gradient of the log-posterior---derived from the Supernet learned distribution---to steer particles toward regions of the search space with high predictive likelihood. The kernel $k(\tilde{x}_j, \tilde{x})$ acts as a weighting function, allowing high-probability particles to influence the trajectory of their neighbors, thereby ensuring that all ensemble members converge toward valid, high-accuracy architectures.

    \item \textbf{Repulsive Force (Diversity $\mathcal{F}_D)$)}:  
    
    The second term, $(1 + \delta) \nabla_{\tilde{x}_j} k(\tilde{x}_j, \tilde{x})$, acts as the exploratory component. Since the gradient of the Radial Basis Function (RBF) kernel, $\nabla_{\tilde{x}_j} k(\tilde{x}_j, \tilde{x})$, points away from the particle locations, this term creates a dispersive vector field that pushes particles apart. 

    \item \textbf{Diversity Coefficient ($\delta$)}: This hyper-parameter artificially amplifies the repulsive force beyond the standard SVGD formulation. 
\end{itemize}

\section{Datasets}
\label{sec:datasets}
We evaluate our methods on three benchmark datasets: CIFAR-10~\cite{cifar}, CIFAR-100~\cite{cifar}, and ImageNet-16-120.
CIFAR-10 and CIFAR-100~\cite{cifar} consist of 32$\times$32 color images, representing 10 and 100 classes, respectively. Each dataset contains 60,000 images, split into 50,000 training images and 10,000 test images. ImageNet-16-120~\cite{imagenet16} is a downsized version of the ImageNet dataset specifically designed for NAS evaluation, consisting of 16$\times$16 color images spanning 120 object classes. It contains 151,700 training images and 6,000 test images.

\section{Different Thresholds and Exit criteria}
\label{sec:exitcriteria}
In this section, we report the value of accuracy and $F_M$ by varying the $\tau_{conf}$ and the exit criteria of SQUAD (mean confidence or t-test) , as explained in section \ref{subsec:tstudent}. In general, t-test leads to solutions with higher accuracy and $F_M$ but, interestingly, the mean confidence seems to perform slightly better on CIFAR-10. 

\begin{figure*}[h!]
    \centering
    \scalebox{0.85}{
        \includegraphics[width=\textwidth]{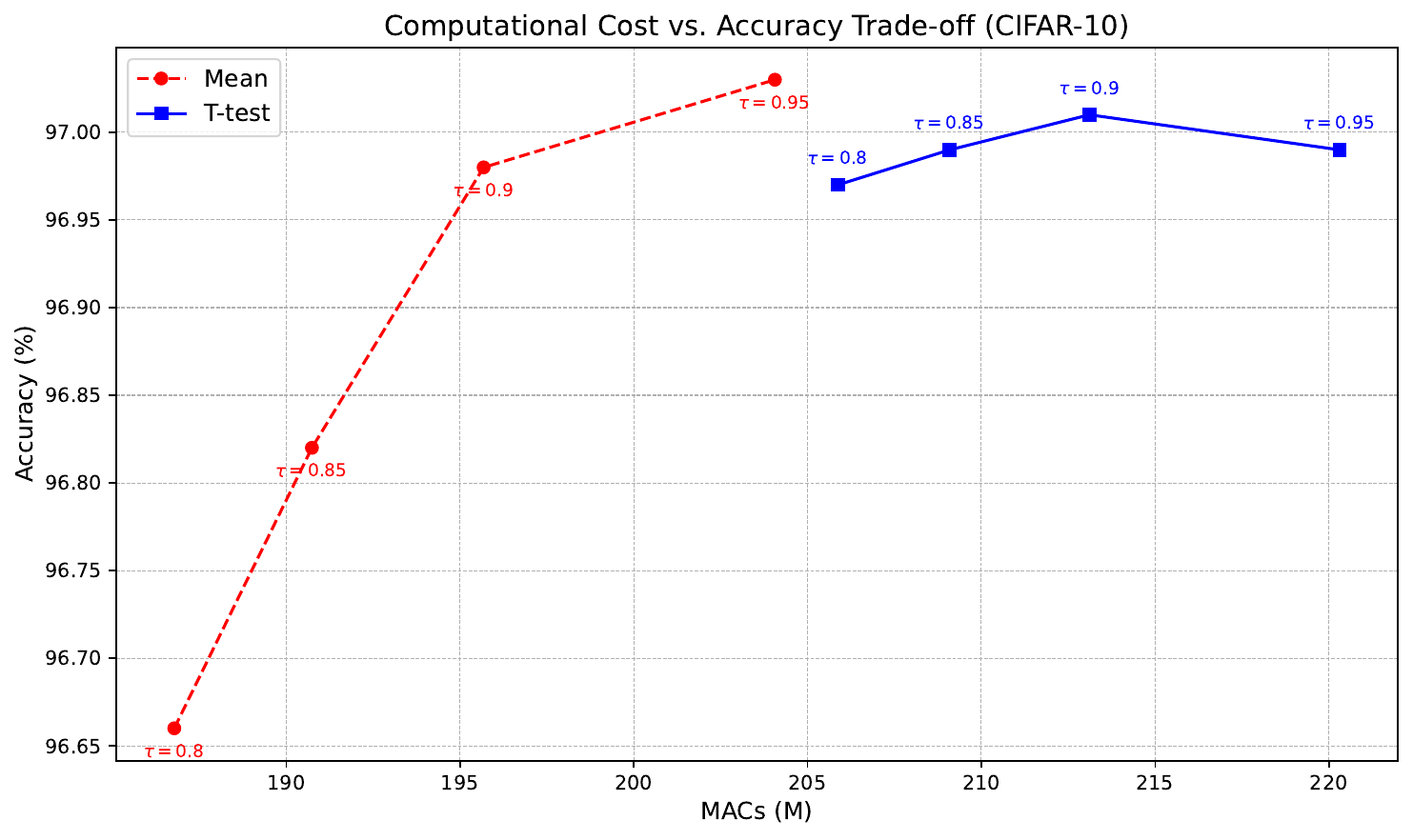}
    }
    \caption{ Mean vs T-student early exit criterion on CIFAR-10 with different $\tau_{conf}$}
    \label{fig:exitcriterioncifar10}
\end{figure*}

\begin{figure*}[h!]
    \centering
    \scalebox{0.85}{
        \includegraphics[width=\textwidth]{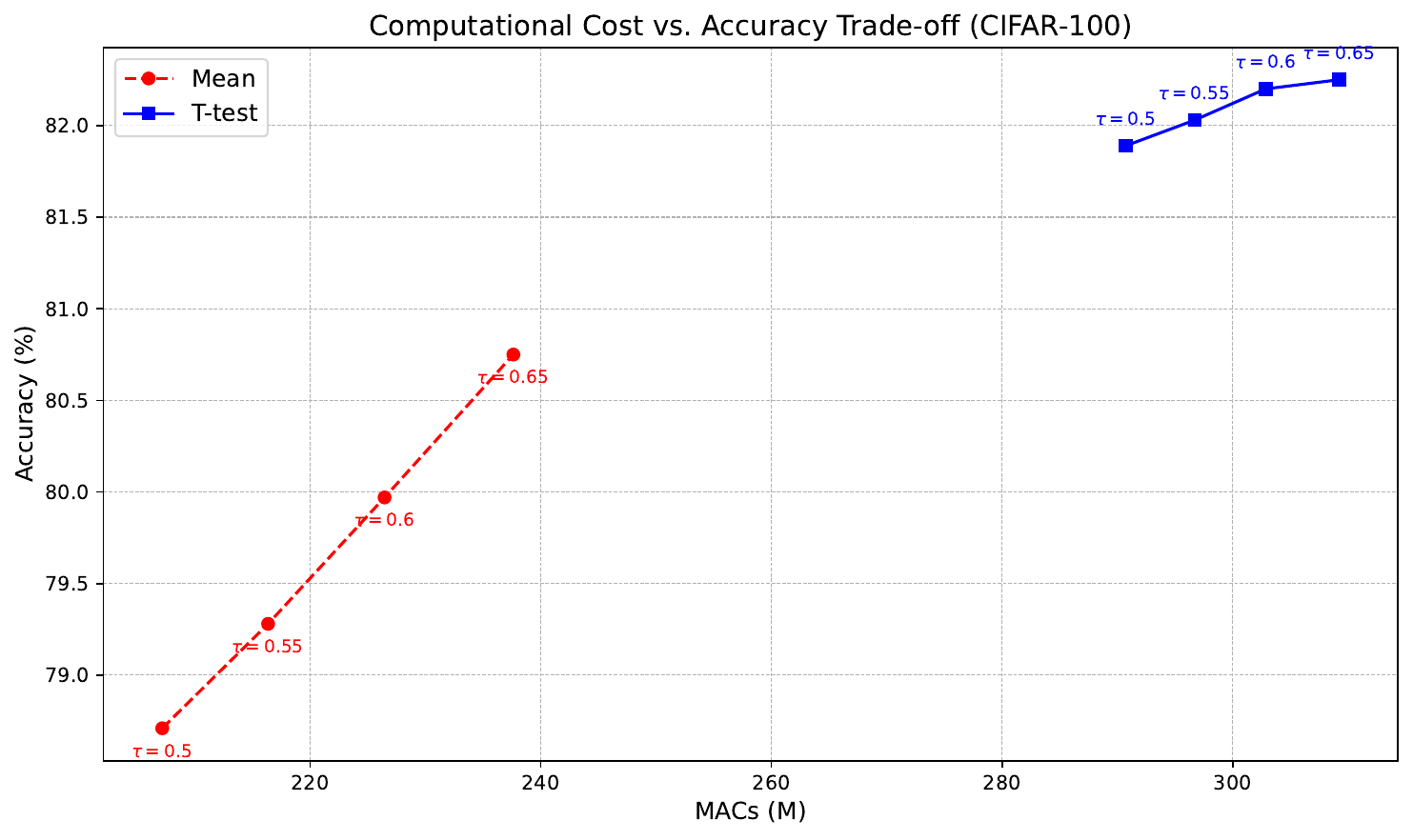}
    }
    \caption{Mean vs T-student early exit criterion on CIFAR-100 with different $\tau_{conf}$}
    \label{fig:exitcriterioncifar100}
\end{figure*}

\begin{figure*}[h!]
    \centering
    \scalebox{0.85}{
        \includegraphics[width=\textwidth]{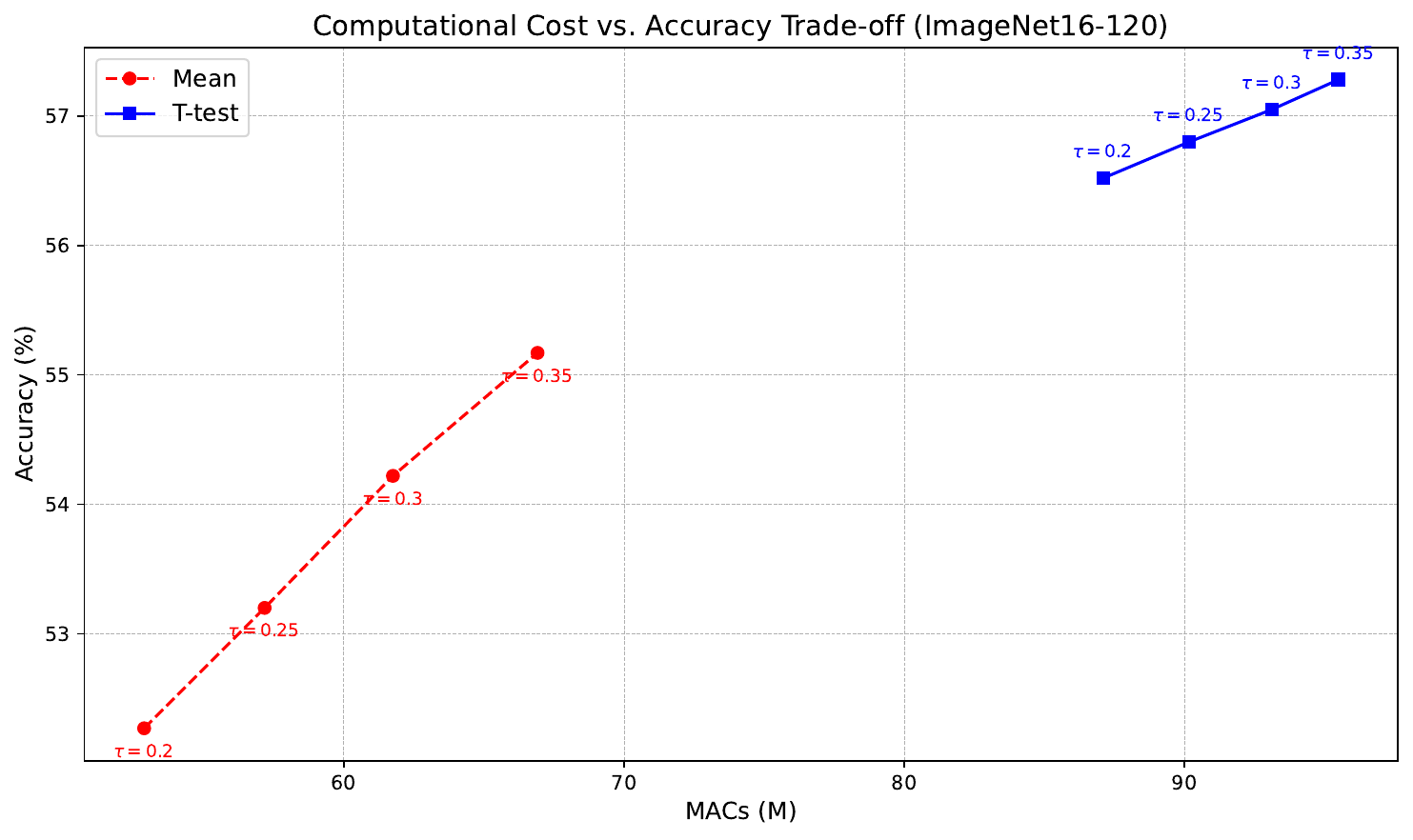}
    }
    \caption{Mean vs T-student early exit criterion on ImageNet16-120 with different $\tau_{conf}$}
    \label{fig:exitcriterionimagenet16}
\end{figure*}
\newpage

\section{Train and search hyperparameters}
\label{sec:hyperparameters}

This section shows the specific hyperparameters of the search and train phase of our NAS on the DARTS search space. The search consists of three steps: supernet training, computation of the posterior distribution, and SVGD-RD search.

\begin{table}[h!]
\centering
\caption{Hyperparameters for Supernet Training on CIFAR-10 on DARTS search space}
\scalebox{1.0}{
\begin{tabular}{|l|l|}
\hline
\textbf{Hyperparameter} & \textbf{Value / Setting} \\
\hline
Dataset & CIFAR-10 (70\% split train) \\
Epochs & 150 \\
Batch size & 64 \\
Optimizer (weights) & SGD (momentum = 0.9, weight decay = $3 \times 10^{-4}$) \\
Optimizer (arch parameters) & Adam (lr = $3 \times 10^{-4}$, betas = (0.5, 0.999), weight decay = $10^{-3}$) \\
Learning rate & decay from 0.025 to 0.001 (cosine annealing) \\
Gradient clipping & 5.0 \\
Initial channels & 16 \\
Number of cells & 8 \\
Weight decay (arch parameters) & $10^{-3}$ \\
Weight decay (model weights) & $3 \times 10^{-4}$ \\
Data augmentation & Random Horizontal Flipping, Random Cropping with Padding, and Normalization \\
Cost penalty (MACs) $\lambda$ & $10^{-3}$ \\
Early exit weights $\beta$ & all equal to 1.0 \\
\hline
\end{tabular}
}
\label{tab:settingsupernetdarts}
\end{table}

\begin{table}[h!]
\centering
\caption{Hyperparameters for the computation of the posterior distribution on CIFAR-10 on DARTS search space}
\scalebox{1.0}{
\begin{tabular}{|l|l|}
\hline
\textbf{Hyperparameter} & \textbf{Value / Setting} \\
\hline
Dataset & CIFAR-10 (30\% split train) \\
Initial channels & 16 \\
Number of cells & 8 \\
Epochs & 20 \\
Optimizer (arch parameters) & Adam (lr = 0.01, betas = (0.9, 0.999), weight decay = 3$\times10^{-4}$) \\
Data augmentation & Random Horizontal Flipping, Random Cropping with Padding, and Normalization \\
Early exit weights $\beta$ & all equal to 1.0 \\
\hline
\end{tabular}
}
\label{tab:settingposteriordarts}
\end{table}

\begin{table}[h!]
\centering
\caption{Hyperparameters for the SVGD-RD search on CIFAR-10 on DARTS search space}
\scalebox{1.0}{
\begin{tabular}{|l|l|}
\hline
\textbf{Hyperparameter} & \textbf{Value / Setting} \\
\hline
Epochs & 1000 \\
Optimizer (arch parameters) & SVGD-RD with $\epsilon$ (lr) = 0.1 and momentum=0.9 \\
Number of ensemble members E & 3 \\
Repulsion parameter $\delta$ & -1.3 \\
\hline
\end{tabular}
}
\label{tab:settingsvgddarts}
\end{table}

\begin{table}[t!]
\centering
\caption{Hyperparameters for Final Training Phase on a dataset on DARTS search space}
\scalebox{1.0}{
\begin{tabular}{|l|l|}
\hline
\textbf{Hyperparameter} & \textbf{Value / Setting} \\
\hline
Dataset & Full training set of the dataset \\
Epochs & 600 \\
Batch size & 96 \\
Optimizer & SGD (momentum = 0.9, weight decay = $3 \times 10^{-4}$) \\
Learning rate & decay from 0.025 to 0.001 (cosine annealing) \\
Gradient clipping & 5.0 \\
Initial channels & 36 \\
Number of cells & 20 \\
Early exit weights $\beta$ & all equal to 1.0 \\
Drop path probability & 0.2 (linearly increased) \\
Cutout & True (length = 16) \\
Weight decay & $3 \times 10^{-4}$ \\
Data augmentation & Random Horizontal Flipping, Random Cropping with Padding, and Normalization \\
\hline
\end{tabular}
}
\label{tab:settingtraindarts}
\end{table}

\end{document}